# MedM2T: A MultiModal Framework for Time-Aware Modeling with Electronic Health Record and Electrocardiogram Data

Yu-Chen Kuo
*Institute of Computer Science and Engineering, National Yang Ming Chiao Tung University*
Hsinchu, Taiwan
yckuo.cs12@nycu.edu.tw

Yi-Ju Tseng
*Department of Computer Science, National Yang Ming Chiao Tung University*
Hsinchu, Taiwan
*Computational Health Informatics Program, Boston Children's Hospital*
Boston, MA, USA
yjtseng@nycu.edu.tw

***Abstract*—The inherent multimodality and heterogeneous temporal structures of medical data pose significant challenges for modeling. We propose MedM2T, a time-aware multimodal framework designed to address these complexities. MedM2T integrates: (i) Sparse Time Series Encoder to flexibly handle irregular and sparse time series, (ii) Hierarchical Time-Aware Fusion to capture both micro- and macro-temporal patterns from multiple dense time series, such as ECGs, and (iii) Bi-Modal Attention to extract cross-modal interactions, which can be extended to any number of modalities. To mitigate granularity gaps between modalities, MedM2T uses modality-specific pre-trained encoders and aligns resulting features within a shared encoder. We evaluated MedM2T on MIMIC-IV and MIMIC-IV-ECG datasets for three tasks that encompass chronic and acute disease dynamics: 90-day cardiovascular disease (CVD) prediction, in-hospital mortality prediction, and ICU length-of-stay (LOS) regression. MedM2T achieved superior or comparable performance relative to state-of-the-art multimodal learning frameworks and existing time series models, achieving an AUROC of 0.932 and an AUPRC of 0.670 for CVD prediction; an AUROC of 0.868 and an AUPRC of 0.470 for mortality prediction; and Mean Absolute Error (MAE) of 2.33 for LOS regression. These results highlight the robustness and broad applicability of MedM2T, positioning it as a promising tool in clinical prediction. We provide the implementation of MedM2T at https://github.com/DHLab-TSENG/MedM2T.**

## I. INTRODUCTION

Medical data is inherently rich in both modality and temporality. Clinical decision-making often relies on integrating longitudinal, multi-source, multimodal information, such as laboratory tests and medical imaging, to form a comprehensive view of patient status. Advances in artificial intelligence have enabled significant progress in exploring multimodal data; however, effectively integrating and extracting latent information remains challenging. Variability and heterogeneity in multimodal data require advanced modeling techniques to address these challenges [1].

Frameworks like HAIM [2] employ early fusion strategies by extracting feature representations from images, text, and structured records through embedding techniques and unifying them as model inputs, achieving success in multiple prediction tasks. MultiBench [3] improves flexibility by providing modular components at various stages of multimodal integration, enabling adaptation to different modalities and tasks. MultiModN [4] addresses non-random missing data by transmitting state information across modalities and estimating modality contributions, improving interpretability and maintaining robustness even when certain modalities are missing. These frameworks demonstrate the potential of multimodal integration to improve model performance and adaptability across various healthcare applications.

Despite these advances, discrepancies in modality granularity persist; fine-grained data (e.g., ECG signals) often require complex models to extract meaningful information, whereas coarse-grained data (e.g., demographic variables) can be effectively represented with simpler approaches. Such differences can lead to inconsistent convergence speeds during training, further complicating multimodal modeling [5]. Another major challenge in multimodal learning is effectively fusing and extracting cross-modal information. Previous research tackled this by proposing an attention-based framework using a shared encoder, highlighting crucial modality and achieving accurate predictions [6].

In addition to modality diversity, temporal and longitudinal characteristics in medical data present unique modeling challenges. Longitudinal medical records, laboratory test results, treatment processes, and others capture patient trajectories. However, temporal data in electronic health records (EHRs) often exhibit sparsity and irregularity [7]. Sparsity arises from infrequent observations, potentially obscuring meaningful patterns and weakening feature robustness, diminishing model performance. Irregularity is reflected in non-fixed observation intervals and substantial

Missing Not at Random (MNAR) data, often linked to patient conditions. For instance, more frequent measurements during deterioration and fewer tests during stability. These complexities impact data completeness and increase the difficulty of modeling [8].

Existing solutions include data imputation, which aims to regularize temporal sequences but may introduce bias under high sparsity. Specialized models such as T-LSTM [9], capture features from irregular intervals to improve robustness. Embedding techniques have also been applied to compress sparse data into low-dimensional vectors, enhancing efficiency. More recent methods, such as STraTS [10], employ triple embeddings and self-attention to capture temporal patterns in sparse and irregular time-series data.

Nevertheless, medical data often exhibit heterogeneous temporal structures across multiple scales: micro-temporal data (e.g., electrocardiogram [ECG] signals) is dense and regular, while macro-temporal data (e.g., longitudinal medical records) is sparse and irregular. Prior studies often focused on a single time scale or attempted to model different scales with a unified approach without fully accounting for their intrinsic heterogeneity [7], [11]. Medformer [12] employs multi-granularity patching with self-attention to capture multi-scale dependencies in EEG/ECG signals, while mainly focusing on consecutive relations.

To address these multimodal and temporal challenges, we propose MedM2T, a multimodal learning framework for handling sparsity, irregularity, and hierarchical temporal characteristics in medical data (**Fig. 1**). For processing heterogeneous temporal data, the framework employs two specialized modules: Sparse Time Series Encoding, which flexibly handles numerical and categorical variables in irregular time series, and Hierarchical Time-Aware Fusion, which captures multi-scale temporal patterns from dense data by integrating micro- and macro-level features. For cross-modal integration, MedM2T incorporates Bi-Modal Attention, an extensible mechanism that extracts latent relationships between any pair of modalities [13]. Finally, to bridge the granularity gap between modalities, the framework leverages modality-specific pre-trained encoders and aligns their feature representations through a shared encoder.

We evaluate MedM2T on three clinical tasks with a diverse set of tabular, time-series, signal, and textual data. The results show that MedM2T improves the performance by effectively extracting meaningful patterns across modalities, thereby demonstrating its robustness and broad applicability for complex medical data analysis.

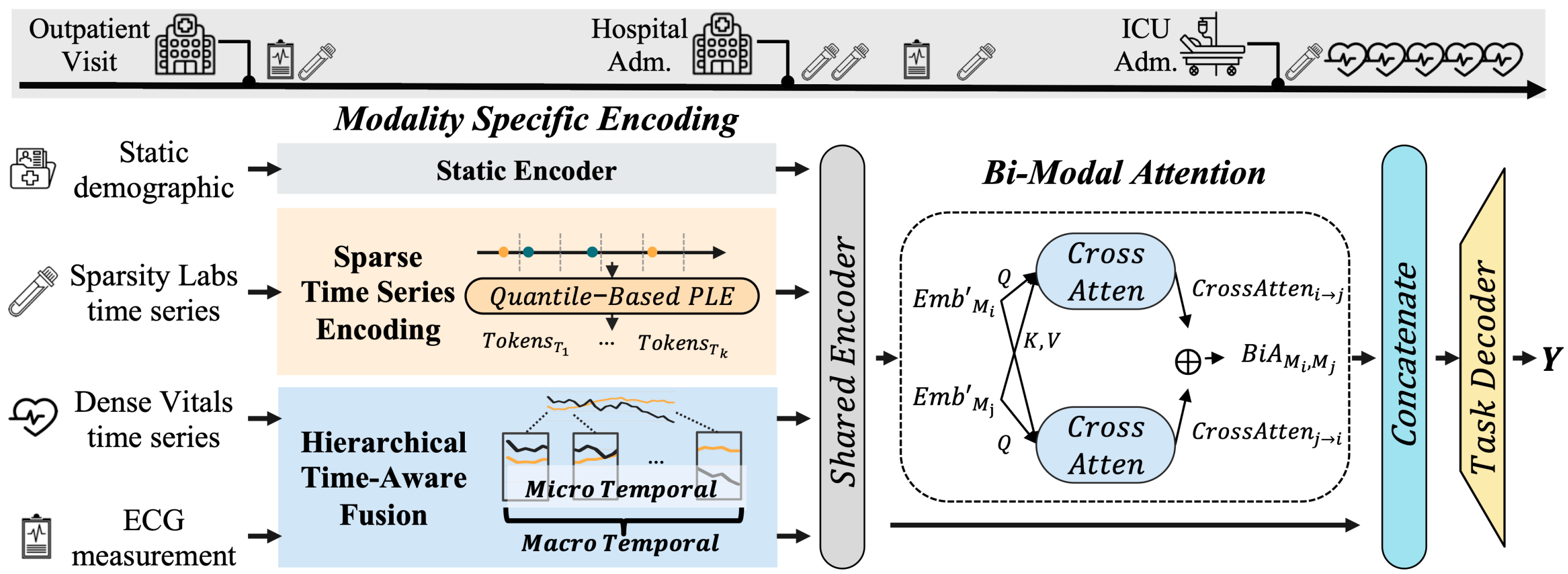

Fig. 1. Overview of the MedM2T framework. Sparse Time Series Encoding and Hierarchical Time-Aware Fusion handle heterogeneous temporal patterns, while modality-specific pre-trained encoders and shared encoder extract and align features. Bi-Modal Attention extracts cross-modal interactions across paired modalities, and task-specific decoders generate predictions.

## II. Methods

### A. Dataset and Evaluation Tasks

We used the MIMIC-IV [14], [15] dataset as the source of EHRs to validate our proposed framework, MedM2T. Three clinical tasks were employed for evaluation: predicting cardiovascular disease (CVD), in-hospital mortality, and length of stay.

#### *MIMIC-IV Dataset*

MIMIC-IV v2.2 provides extensive EHRs sourced from intensive care units (ICU), hospitalizations, and outpatient encounters. ICU records provide dense, short-term data capturing rapid disease progression, while hospitalization and outpatient data are relatively sparse and irregular, reflecting longer-term trends. The multimodal nature of this dataset, which includes tabular, time-series, signal, and text data, facilitates robust evaluations of framework adaptability and performance across diverse clinical scenarios [14], [15], [16].

#### *Task 1: Cardiovascular Disease (CVD) Prediction*

The first task is 90-day prediction of CVD-related hospitalization, defined as a hospitalization where the primary discharge diagnosis is CVD or with CVD-related operation, and identification of specific types of CVD, including coronary artery disease, stroke, and heart failure. By leveraging longitudinal data spanning months to years, the model provides early warnings to guide timely interventions.

The dataset integrates records from MIMIC-IV and MIMIC-IV-ECG [17] modules, focusing on patients with at least one hospitalization occurring within 90 days after an ECG measurement. Patients were excluded if they were under 18 or over 89 years of age, or if their hospital stays were shorter than 24 hours. Since a patient may have multiple ECGs, each ECG record is treated as an independent sample. Each sample was labeled according to the subsequent hospitalization outcome: a 'CVD' label and the type of CVD were assigned if the subsequent admission was CVD-related, and 'non-CVD' otherwise. The final dataset comprises 125,987 non-CVD and 44,790 CVD samples. To prevent data leakage, the observation window for feature extraction included all historical data up to a cutoff point defined as the earlier of two events: three days after the index ECG measurement, or the day prior to CVD-related hospitalization. Key multimodal features include:

1) **EHR Static Data:** Patient demographics, latest outpatient measurements (e.g., blood pressure), and binary labels for pre-existing CVD-related medical history and medications.
2) **Sparse Laboratory Results:** Eight CVD-related laboratory tests with irregular time-series characteristics.
3) **ECG Signals, Text, and Features:** Standard 12-lead ECG signals were down-sampled to 125Hz, with 5-second segments extracted for efficient processing. Machine-generated reports include clinical notes and time-domain features such as heart rate and PR intervals. Clinical notes underwent additional preprocessing and were mapped to 143 SNOMED CT clinical terms, offering structured and interpretable diagnostic judgments.

***Task 2: In-hospital Mortality Prediction***

The second task involves predicting in-hospital mortality based on data from the first 24 hours of a patient's first ICU admission, addressing the need for timely decisions in critical care. Records were excluded if their ICU stays were shorter than 24 hours. The dataset contains 40,167 first ICU admission records, including 4,035 mortality cases. Key multimodal features include:

1) **EHR Static Data:** Patient demographics and admission details, such as age and admission type.
2) **Dense Vital Signs**: Hourly measurements of 24 time-series variables are included, excluding those with over 80% missing values.
3) **Sparse Laboratory Results**: 74 laboratory tests are included as time series data, excluding those with over 80% missing values.
4) **ECG Signals, Text, and Features**: Same as Task 1.

***Task 3: Length of Stay (LOS)***

The third task is predicting the length of ICU stays, using the same dataset from Task 2 to assess model performance in a regression task.

These 3 tasks highlight the challenges of sparse and irregular time-series data, emphasizing the model's capacity for managing multimodal information effectively. Detailed procedures for population selection and dataset descriptions are provided in the Supplementary Material (A: Dataset) [34].

### *B. Time-Aware Modeling: Sparse Time Series Encoding*

The Sparse Time Series Encoding module is designed to handle sparse and irregular time-series data by converting them into a sequence of embeddings, thereby bypassing imputation which often introduces noise or bias [8] (**Fig. 2**). Inspired by natural language processing, numerical values are discretized into value tokens using a quantile-based piecewise linear encoding (quantile-based PLE) tokenizer [18]. This approach preserves the relative magnitudes of original data while ensuring a balanced distribution across the token vocabulary. Then, source tokens are added to label the origin of each value, enhancing contextual understanding.

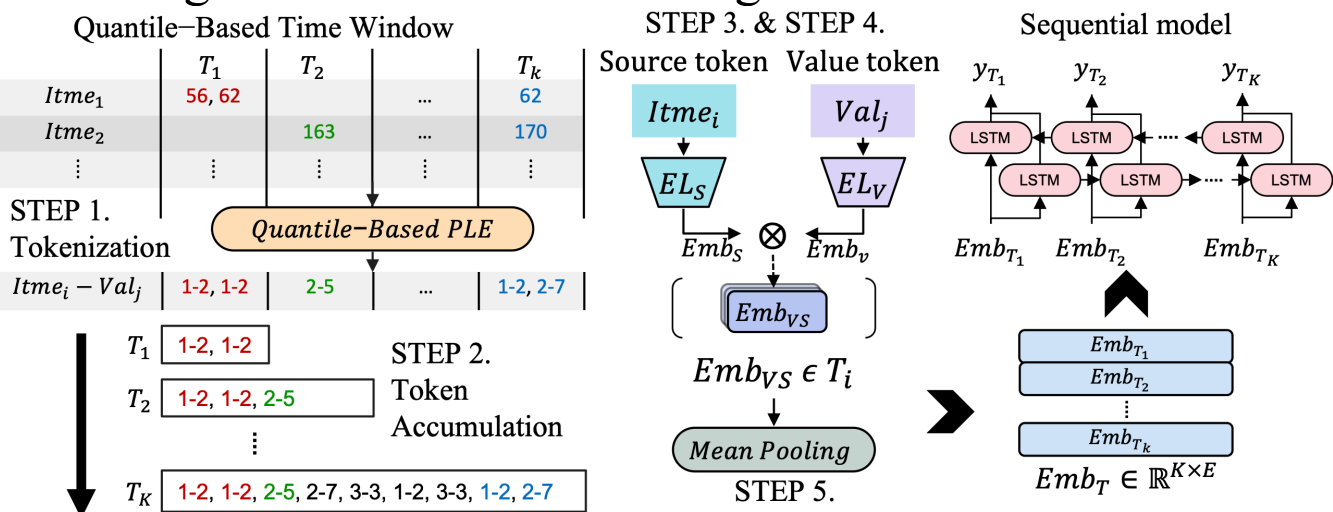

Fig. 2. Processing pipeline of Sparse Time Series Encoding using quantile-based PLE and quantile-based time window embedding. Raw numerical values are tokenized based on quantiles, accumulated within time windows, and transformed into embeddings. Mean pooling aggregates embeddings per window, which are fed into a sequential model to capture temporal patterns for prediction. **EL:** Embedding Layer, **Emb:** Embedding, **V:** Values, **S:** Sources, **T:** Time Window.

To handle irregular time intervals, we applied a quantile-based time window embedding strategy. This approach partitions the timeline into non-uniform, data-driven windows defined by the quantiles of time. In each window, a token accumulation aggregates multiple values within the same time window and prior windows into a single representation without discarding any records. It preserves essential temporal and contextual information while minimizing the impact of sparsity.

We adapt the time-slicing strategy based on the data distribution specific to each task. The CVD prediction (Task 1) model uses a biased slicing approach to prioritize time windows closer to the ECG measurement, reflecting the fact that critical laboratory data are often cluster near the time of diagnosis. In contrast, the in-hospital mortality prediction (Task 2) model adopts uniform slicing, as the time-series data is relatively dense within the 24-hour window before ICU admission. Time windows are defined relative to key timestamps:

1) **Task 1 (CVD prediction):** $\Delta t = t_{ecg} - t_{event}$, where time bins are defined at the {0th, 5th, 10th, 20th, 40th, 80th, 100th} percentiles.
2) **Task 2 (mortality prediction) and Task 3 (LOS prediction):** $\Delta t = t_{event} - t_{in}$, with time bins spanning the {0th, 10th, 20th, ..., 100th} percentiles.

where $t_{event}$ is the event timestamp (e.g., lab test), $t_{ecg}$ is the latest ECG timestamp, and $t_{in}$ is ICU admission timestamp.

Within each time window, embeddings are computed through the following steps (**Fig. 2**):

1) **Tokenization**: Transform values and sources into tokens ($Val$ and $Item$), with numerical values processed using the Quantile-Based PLE.
2) **Token Accumulation**: Aggregate multiple tokens within the same and prior time windows $T_{i \leq k}$.

3) **Embedding Generation**: Convert value and source tokens into value $Emb_v$ and source $Emb_s$ embeddings.
4) **Embedding Combination**: Combine value and source embeddings using element-wise multiplication, generating $Emb_{VS}$ for each time window.
5) **Pooling**: Apply mean pooling to create a single embedding $Emb_{T_i}$ representing the time window $T_i$.

The aggregated embeddings from all-time windows are processed by a bidirectional long short-term memory (BiLSTM) model to capture dynamic temporal patterns, enabling robust modeling of sparse and irregular time-series data.

### *C. Time-Aware Modeling: Hierarchical Time-aware Fusion*

The Hierarchical Time-aware Fusion framework is designed to capture the multi-scale nature of temporal patterns in dense medical data, such as signal and vital signs. The framework includes high-frequency encoders to capture fine-grained, short-term variations, and low-frequency encoders to identify coarse-grained, long-term trends. The framework constructs a comprehensive representation that encompasses the full spectrum of temporal dynamics.

#### ***Hierarchical Time-Aware Fusion Model for ECG Data***

ECG data consists of three modalities:

1) **Signal Modality**: ECG signals represent millisecond-level micro-temporal changes processed using a ResNet-based high-frequency encoder.
2) **Text Modality**: Machine-generated text reports are mapped to SNOMED CT clinical terms, tokenized, and processed through an embedding layer to generate text embeddings.
3) **Feature Modality**: ECG time domain features are processed via a multilayer perceptron (MLP) to extract feature embeddings.

We use a two-level hierarchical framework with ResNet-based encoders, inspired by an architecture for ECG diagnosis [19]. First, the high-frequency encoder extracts micro-temporal embeddings from ECG signals, capturing short-term variations. Subsequently, these embeddings are fused with text and feature embeddings to generate comprehensive ECG-level representations for each ECG. To address the sparsity and irregularity inherent in multiple ECG, the timeline is partitioned using quantile-based windows, and the representations within each window are aggregated through mean pooling. The aggregated embeddings from each time window are then processed by the low-frequency encoder to extract macro-temporal embeddings that capture long-term trends (**Fig. 3**).

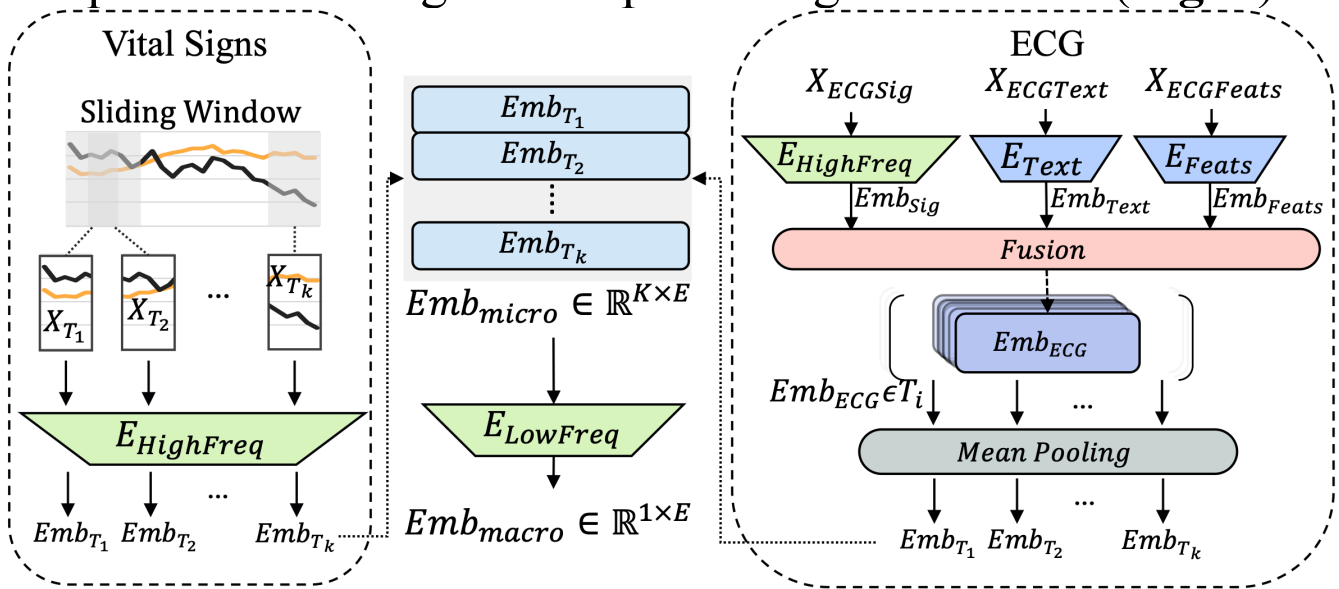

Fig. 3. Hierarchical Time-Aware Fusion Model for vital Signs and ECG data. For ECG data, high-frequency encoders generate embeddings from signals. The signal, features and text embeddings from the same measurement are fused and then pooled within time windows. Followed by a low-frequency encoder to get macro-temporal representations. For vital signs, sliding windows segment the time series, and high-frequency encoders extract micro-temporal embeddings, which are then aggregated by a low-frequency encoder to model long-term trends. The model captures both multimodal and multiscale temporal dependencies. **E**: Encoder, **Emb**: Embedding, **HF**: High-Frequency, **LF**: Low-Frequency, **T**: Time Window.

#### ***Hierarchical Time-Aware Fusion Model for Vital Signs***

Vital signs, which are typically dense and regularly monitored, especially in ICU, can be considered continuous physiological signals. We use a sliding window approach to extract features across micro- and macro-temporal scales in vital signs. The following are the main steps (**Fig. 3**):

1) **Data Preprocessing and Upsampling**: Originally recorded hourly, vital signs are upsampled to 15-minute intervals to increase data points and aligned the time with different items. Missing data is imputed using linear interpolation when both previous and subsequent observations exist, nearest-neighbor interpolation when only one adjacent observation is available, and zero imputation for entirely missing values.
2) **Normalization**: Z-score normalization applies to address variations between vital sign values.
3) **Sliding Window Mechanism**: Time series are segmented into overlapping windows (1) of size W minutes with a step size S.

$$window_k = \{x_t | t \in [t_0 + (k-1)S, t_0 + (k-1)S + W]\}. \quad (1)$$

4) **High-Frequency Encoding**: Self-attention mechanisms capture interrelations among vital signs at each time point [13], while ResNet-based high-frequency encoder extracts micro-temporal features from each window.
5) **Low-Frequency Encoding**: Sequential window embeddings are processed through a ResNet-based low-frequency encoder to extract macro-temporal patterns.

Multi-scale feature extraction ensures robustness across varying temporal resolutions.

### *D. Multi-Modality Framework*

To address challenges in multimodal learning, including capturing cross-modal interactions and handling convergence inconsistencies among modalities, we propose a flexible and scalable fusion framework which is adaptable to any number of modalities. The framework ensures efficient and robust learning across modalities (**Fig. 4**).

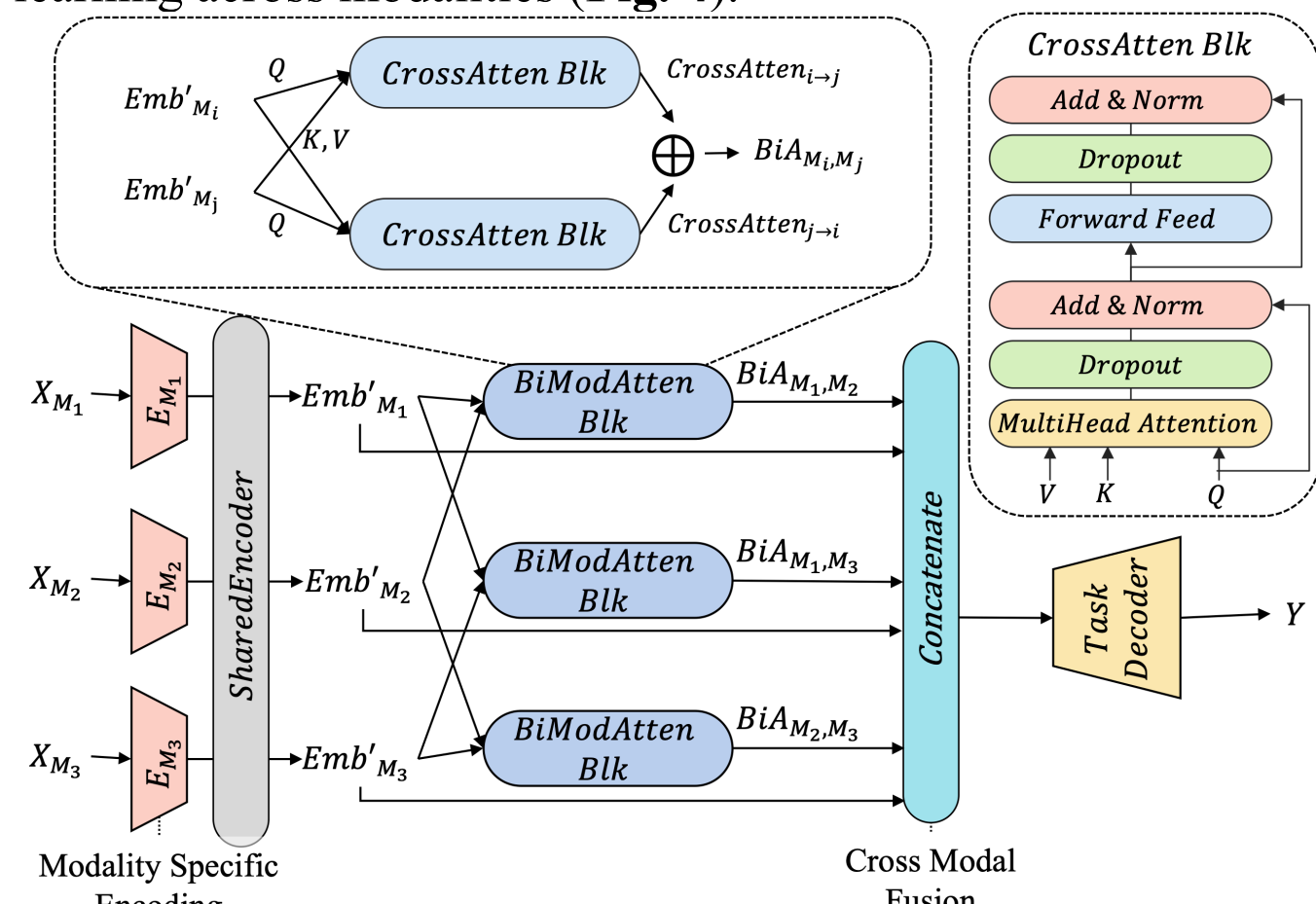

Fig. 4. Overview of the Multi-Modality Framework. The framework processes input modalities using modality-specific encoders and aligns them through a shared encoder. Bi-Modal Attention (**BiA**) blocks capture cross-modal interactions, while embeddings and attention outputs are fused into a unified representation for task decoding. **E:** Encoder, **Emb:** Embedding, **M:** Modality.

### *Modality-Specific Encoding*

We process each data modality with a specialized encoder architecture: time-aware models, as described in the previous session, for temporal data like laboratory results or ECG signals, and MLPs for static data. Given a set of modalities $\mathcal{M} = \{M_1, M_2, \ldots, M_N\}$, each raw input $X_{M_i}$is used to pre-train these modality-specific encoders (**Algorithm 1**). This pre-training step reduces convergence disparities and improves efficiency [5]. The pre-trained encoders are then used to extract latent features, as shown in **Algorithm 2** (1.1), with each encoder customized to the unique characteristics of its respective modality to optimize feature extraction.

### *Shared Encoding*

To align embeddings across modalities, a shared encoder projects modality-specific embeddings into a unified space. The shared layer aligns these transformed embeddings into a common space, as defined in **Algorithm 2** (2.1, 2.2). The aligned embeddings facilitate the extraction of cross-modal interactions in subsequent stages [6].

### *Bi-Modal Attention*

The Bi-Modal Attention mechanism models interactions between pairs of modalities $M_i$ and $M_j$ . For each pair, bidirectional attention $\text{BiAtten}_{M_i,M_j}$ aggregates features, as defined in **Algorithm 2** (3.3), where $\text{CrossAtten}_{i\rightarrow j}$ computers attention from $M_i$ to $M_j$, through the aligned embeddings, as shown in **Algorithm 2** (3.1).

This mechanism, inspired by prior works [20], [13], is designed to flexibly accommodate an arbitrary number of modalities. The bidirectional architecture mitigates over-reliance on a single modality by balancing attention across paired modalities. This design enables robust cross-modal feature extraction, fostering the comprehensive integration of information from diverse modalities.

### *Cross-Modal Fusion*

All modality embeddings and their Bi-Modal Attention results are concatenated into a unified representation, as defined in **Algorithm 2** (4.1). This approach preserves intra-modality features while capturing inter-modality interactions, enabling robust multimodal learning.

### *Task Decoding*

The unified cross-modal embedding is fed into task-specific decoders for classification or regression. For classification tasks in this study, MLP decoders with LogSoftmax activation produce the final predictions, as shown in **Algorithm 2** (5.1).

By efficiently integrating intra- and inter-modality features, this framework enhances model performance across diverse multimodal tasks.

**Algorithm 1:** Modality-Specific Encoder Pretraining

**Input:** Modality set $M = \{M_1, M_2, \ldots, M_N\}$, Target $T$
**Output:** Pretrained modality-specific encoders $\text{Encoder}_{M_i}$ for each $M_i$

1 **foreach** $M_i \in M$ **do**
2   Initialize $\text{Encoder}_{M_i}$ and $\text{TaskDecoder}_{M_i}$
3   **for** *epoch = 1* ***to*** *MaxEpoch* **do**
4     $\text{Embed}_{M_i} \leftarrow \text{Encoder}_{M_i}(M_i)$
5     $\text{Prediction} \leftarrow \text{TaskDecoder}_{M_i}(\text{Embed}_{M_i})$
6     $\text{Loss} \leftarrow \text{LossFunction}(\text{Prediction}, T)$
7     Backpropagation(Loss)
8   Save $\text{Encoder}_{M_i}$ with minimum validation loss
9 **return** $\{Encoder_{M_i} \mid M_i \in M\}$

**Algorithm 2:** Multi-Modality Framework

**Input:** Modality set $M = \{M_1, M_2, \ldots, M_N\}$, Target $T$
**Output:** Task-specific predictions

1 **Step 1: Modality-Specific Encoding**
2 **foreach** $M_i \in M$ **do**
3   $\text{Embed}_{M_i} \leftarrow \text{Encoder}_{M_i}(X_{M_i})$ (1.1)
  `// Encoder`$_{M_i}$ `is pretrained`
4 **Step 2: Shared Encoding**
5 **foreach** $Embed_{M_i}$ **do**
6   $\text{Embed}'_{M_i} \leftarrow W_{M_i} \cdot \text{Embed}_{M_i} + b_{M_i}$ (2.1)
7   $\text{Embed}'_{M_i} \leftarrow W_{\text{shared}} \cdot \text{Embed}'_{M_i} + b_{\text{shared}}$ (2.2)
8 **Step 3: Bi-Modal Attention**
9 **foreach** $(M_i, M_j) \mid i \neq j$ **do**
10   $\text{CrossAtten}_{i\rightarrow j} \leftarrow \text{Attention}(\text{Embed}'_{M_i}, \text{Embed}'_{M_j}, \text{Embed}'_{M_j})$ (3.1)
11   $\text{CrossAtten}_{j\rightarrow i} \leftarrow \text{Attention}(\text{Embed}'_{M_j}, \text{Embed}'_{M_i}, \text{Embed}'_{M_i})$ (3.2)
12   $\text{BiAtten}_{M_i,M_j} \leftarrow \text{CrossAtten}_{i\rightarrow j} + \text{CrossAtten}_{j\rightarrow i}$ (3.3)
13 **Step 4: Cross-Modal Fusion**
14   $\text{CrossModalEmbed} \leftarrow [\text{Embed}'_{M_i} \mid i = 1, 2, \ldots, N]$
  $\cup [\text{BiAtten}_{M_i,M_j} \mid i = 1, 2, \ldots, N; j = i + 1, \ldots, N]$ (4.1)
15 **Step 5: Task Decoding**
16   $\text{Output} \leftarrow \text{TaskDecoder}(\text{CrossModalEmbed})$ (5.1)
17 **return** *Output*

## III. RESULTS

### *A. Experiments Setup*

The proposed approach, MedM2T, was evaluated across three clinical tasks: multi-class classification (Task 1), binary classification (Task 2), and regression (Task 3), each reflecting distinct temporal patterns. A five-fold cross-validation was applied, with training, validation, and test sets split in a 0.64:0.16:0.20 ratio. To avoid data leakage, cross-subject partitioning was used. Classification models were trained with cross-entropy loss and evaluated by AUROC and AUPRC, with macro-averaging applied in multi-class. For regression, MSE was used as the training loss, with both MSE and MAE reported. An early stopping strategy was applied during training, and the model checkpoint with the lowest validation loss was selected for testing.

For unimodal experiments, training and evaluation were conducted only on the subset of data available for that specific modality. In contrast, results with two or more modalities were obtained using the complete data.

The best-performing results are highlighted in bold. Permutation tests were implemented for classification and paired t-tests for regression. Results that are not statistically significantly different from the best ($\alpha = 0.05$) are also marked in bold, indicating comparable performance.

Detailed configurations of the model hyperparameters, along with the sample sizes for each task and modality, are provided in the Supplementary Material (B: Experimental Setup) [34].

### B. Evaluation of Time-Aware Modeling

#### *Sparse Time Series Encoding*

**Table I** presents the performance of our Sparse Time Series Encoding relative to various benchmarks and state-of-the-art method for irregular clinical data (STraTS [10]). The baseline uses only the single most recent record. LSTM and GRU employ a quantile-based windowing strategy, consistent with our methodology, while their inputs were processed using standard forward fill and zero fill imputation.

The proposed method achieved the best performance in Task 1 and the second highest in Task 2. In Task 3, it showed equivalent performance to STraTS.

We further examined the computational cost of STraTS relative to the proposed method across tasks. In Task 1, the proposed method utilizes only 118 MB of reserved GPU memory during training, representing a sharp contrast to the 18.7 GB reserved by STraTS. In terms of per-batch runtime, our approach completes training in 2.9 ms, STraTS requires 21.1 ms under the same settings. In the remaining tasks, STraTS continues to demand significantly higher resources, requiring 7 to 13 GB more GPU memory during training and 3 to 7 GB more during inference than our approach, along with roughly 1.5–2 times longer training duration.

TABLE I

PERFORMANCE COMPARISON OF SPARSE TIME SERIES ENCODING WITH CONVENTIONAL METHODS

| | **Task 1: Labs** | **Task 2: Labs** | **Task 3: Labs** |
|---|---|---|---|
| ***Metric*** | *AUROC / AUPRC* ↑ | | *MAE / MSE* ↓ |
| **Proposed** | **0.870 / 0.568** | <u>0.830 / 0.390</u> | **2.51 / 22.82** |
| **Baseline** | 0.750 / 0.433 | 0.809 / 0.363 | 2.68 / 27.56 |
| **LSTM** | <u>0.839</u> / 0.516 | 0.784 / 0.311 | <u>2.61 / 24.37</u> |
| **GRU** | <u>0.839 / 0.521</u> | 0.780 / 0.313 | 2.70 / 24.44 |
| **STraTS** | 0.813 / 0.502 | **0.845 / 0.443** | **2.50 / 22.68** |

Bold and underlined values represent the best (and statistically comparable, $p< 0.05$) and second-best results, respectively; Proposed denotes Sparse Time Series Encoding; Baseline uses the single most recent record.

#### *Hierarchical Time-Aware Fusion*

**Table II** compares our Hierarchical Time-Aware Fusion method with baselines, which are defined as using the latest ECG record or raw time-series vital signs without hierarchical processing. Results show that our method outperforms the baseline in all vital signs involved tasks and long-term Task 1 with ECG data. In contrast, on short-term ECG tasks (Task 2 and 3), where most patients have only a single ECG record in the 24-hour pre-ICU window, our method is slightly attenuated compared to the baseline. This marginal reduction may be because the core advantage of our hierarchical model, capturing long-term trends, is less effective when longitudinal data is absent.

TABLE II

PERFORMANCE OF HIERARCHICAL TIME-AWARE FUSION

| | ***Metric*** | **Proposed** | **Baseline** |
|---|---|---|---|
| **Task 1: ECG** | *AUROC / AUPRC* ↑ | **0.844 / 0.527** | 0.810 / 0.481 |
| **Task 2: ECG** | | 0.701 / 0.190 | **0.737 / 0.218** |
| **Task 2: Vitals (N)*** | | **0.814 / 0.380** | 0.781 / 0.333 |
| **Task 3: ECG** | *MAE / MSE* ↓ | 2.79 / **26.03** | **2.71 / 26.05** |
| **Task 3: Vitals (N)*** | | **2.50 / 22.47** | 2.57 / 23.39 |

Bold denotes best/comparable results ($p < 0.05$); Vitals (N) denotes numerical vital signs; Proposed denotes Hierarchical Time-Aware Fusion; Baseline uses the latest ECG record or raw vitals without segmentation.

#### *Advanced Time Series Methods*

**Table III** compares our proposed time-aware models with advanced time-series models, all benchmarked on MIMIC mortality prediction tasks [21], [22], [23], [24]. Mortality prediction was chosen for comparison as it is a common and representative benchmark, consistently used by established models to evaluate model performance. Following the established methodology in these studies, we preprocessed the first 24 hours of ICU data by segmenting it into hourly windows, averaging observations, imputing missing values using the forward-fill method, and replacing any remaining missing features with their training-set median. Our proposed method showed consistent and superior performance across most metrics and modalities, highlighting its overall robustness.

TABLE III

COMPARISON WITH ADVANCED TIME-SERIES MODELS FOR MORTALITY PREDICTION

| | **Labs** | **Vitals** | **Vitals + Labs** |
|---|---|---|---|
| ***Metric*** | | *AUROC / AUPRC* ↑ | |
| **Proposed** | **0.830 / 0.390** | **0.818 / 0.387** | **0.860 / 0.458** |
| **BoXHED** | 0.798 / 0.367 | 0.745 / 0.286 | 0.826 / 0.406 |
| **Bi-LSTM** | 0.761 / 0.296 | 0.766 / 0.311 | 0.780 / 0.310 |
| **IMV-LSTM** | 0.813 / **0.396** | 0.751 / 0.295 | 0.828 / 0.419 |

Bold denotes best/comparable results ($p < 0.05$).

TABLE IV

UNIMODAL VS. MULTIMODAL PERFORMANCE ACROSS THREE CLINICAL TASKS

| | | **Task 1: CVD** | | **Task 2: Mortality** | | **Task 3: LOS** | |
|---|---|---|---|---|---|---|---|
| ***Metric*** | | *AUROC* ↑ | *AUPRC* ↑ | *AUROC* ↑ | *AUPRC* ↑ | *MAE* ↓ | *MSE* ↓ |
| **Unimodal** | **Static** | 0.717 / 0.846* | 0.362 / 0.546* | 0.678 | 0.170 | 2.89 | 26.53 |
| | **Labs** | 0.870 | 0.568 | 0.830 | 0.390 | 2.51 | 22.82 |
| | **Vitals** | - | - | 0.818 | 0.387 | 2.51 | 22.66 |
| | **ECG** | 0.844 | 0.527 | 0.710 | 0.191 | 2.79 | 26.03 |
| **Multimodal (proposed)** | | **0.915 / 0.932*** | **0.632 / 0.670*** | **0.868** | **0.470** | **2.33** | **22.21** |

Bold denotes best/comparable results ($p < 0.05$); asterisk (*) denotes using the core subset / extended subset. Additional results are provided in **Supplementary Table I**.

TABLE V

ABLATION STUDY AND COMPARISON OF MEDM2T WITH OTHER MULTIMODAL FRAMEWORKS

| | **Task 1 (core)** | **Task 1 (extended)** | **Task 2** | **Task 3** |
|---|---|---|---|---|

| *Metric* | | *AUROC / AUPRC* ↑ | | | *MAE / MSE* ↓ |
|---|---|---|---|---|---|
| **MedM2T** | | **0.915** / **0.632** | **0.932** / 0.670 | 0.868 / 0.470 | **2.33 / 22.21** |
| **MedM2T (Ablation study)** | **w/o Pre-trained Encoder** | 0.901 / 0.604 | 0.917 / 0.653 | 0.868 / 0.477 | 2.54 / **21.95** |
| | **w/o Bi-Modal Attention** | 0.902 / 0.624 | 0.925 / 0.665 | 0.868 / 0.476 | 2.35 / 22.55 |
| | **w/o Shared Encoder** | 0.909 / **0.628** | **0.930** / 0.666 | 0.868 / 0.472 | 2.36 / **21.91** |
| **MultiBench** | **LF** | 0.895 / 0.617 | 0.915 / 0.649 | 0.833 / 0.418 | 2.93 / 26.82 |
| | **LRTF** | 0.897 / 0.621 | 0.916 / 0.649 | 0.823 / 0.368 | 2.96 / 26.86 |
| **MultiBench (Our Encoder)** | **LF** | 0.890 / 0.583 | 0.923 / 0.661 | 0.862 / 0.460 | 2.96 / 26.88 |
| | **LRTF** | 0.896 / 0.613 | 0.917 / 0.643 | 0.768 / 0.343 | 2.95 / 26.92 |
| **MultiModN** | | 0.871 / 0.573 | 0.889 / 0.593 | 0.856 / 0.409 | 2.95 / 26.92 |
| **MultiModN (Our Encoder)** | | 0.894 / 0.600 | 0.911 / 0.631 | 0.867 / 0.455 | 2.95 / 26.92 |
| **HAIM** | | 0.899 / **0.633** | 0.923 / **0.680** | **0.890 / 0.540** | 2.43 / **21.89** |

Bold and underlined values represent the best (and statistically comparable, $p < 0.05$) and second-best results, respectively; "Our Encoder" refers to substituting their encoders with the modality-specific encoders (w/o pre-trained) used in MedM2T; LF, LRTF denote late fusion and low rank tensor fusion; since MultiBench and MultiModN do not provide recommended encoder for physiological signals, we adopted a ResNet-based encoder and used the latest ECG signal as input. Additional results are provided in **Supplementary Table II**.

TABLE VI
ABLATION RESULTS ON THE ECG-AVAILABLE CICU SUBSET

| *Metric* | Task 2<br>*AUROC / AUPRC*↑ | Task 3<br>*MAE / MSE*↓ |
|---|---|---|
| **Proposed** | **0.885** / 0.332 | **1.79 / 15.35** |
| **w/o Pre-trained Encoder** | 0.873 / **0.385** | 1.99 / 15.75 |
| **w/o BiModal Attention** | 0.876 / 0.327 | 2.03 / 17.11 |
| **w/o Shared Encoder** | **0.886** / 0.324 | **1.79 / 15.25** |

Bold denotes best/comparable results ($p < 0.05$). Proposed denotes MedM2T.

### *C. Evaluation of the Multimodal Framework*

**Table IV** compares unimodal versus multimodal performance across 3 clinical tasks. For Task 1, we evaluate two tiers of static data: the core subset comprising demographics and recent outpatient measurements, and the extended subset, which also includes medical and medication histories. Across all tasks, the results consistently demonstrate that integrating multiple modalities yields significant performance gains over relying on any single data source.

**Table V** presents an ablation study of MedM2T fusion framework and a comparative evaluation against representative multimodal methods [2], [3], [4]. To ensure a fair comparison, the evaluation was conducted under two settings. First, each benchmark model was tested using its original encoders. Second, to specifically assess the fusion component, we tested their performance using our proposed encoder as feature extraction backbone.

The ablation study shows that pre-training the modality-specific encoders is the most critical factor for model performance. Removing this pre-training step caused the most significant performance degradation, particularly for Task 1, highlighting its importance in mitigating convergence disparities across heterogeneous modalities. Removing the Bi-Modal Attention yields a statistically significant but smaller performance drop in most tasks, which proves cross-modal interaction modeling provides additional gains. Finally, removing the Shared Encoder causes minor changes across tasks, suggesting that the fusion pipeline remains reasonably robust even without this alignment module. To further clarify when the proposed fusion mechanism is most beneficial, **Table VI** reports results on an ECG-available CICU (cardiac intensive care unit) subset for Tasks 2 and 3. Due to the higher patient homogeneity within this cohort, the Bi-Modal Attention yields even more performance improvements.

Compared to other multimodal methods, MedM2T achieves strong and often competitive performance across tasks, with clearer advantages in Task 1 and Task 3, while achieving the second-best results in Task 2, where it is only surpassed by HAIM [2]. To further assess the contribution of our encoder architecture, we tested its integration with other fusion methods. This substitution yielded significant performance gains in the MultiModN framework. However, similar integration with MultiBench did not produce consistent improvements.

We conducted ablation experiments on multiclass Task 1 (core subset) to analyze each modality's contribution (**Fig. 5**). The ablation of laboratory data most severely impacted the macro-average, non-CVD, CHD, and stroke prediction accuracy, whereas HF prediction was most degraded by removing the ECG modality.

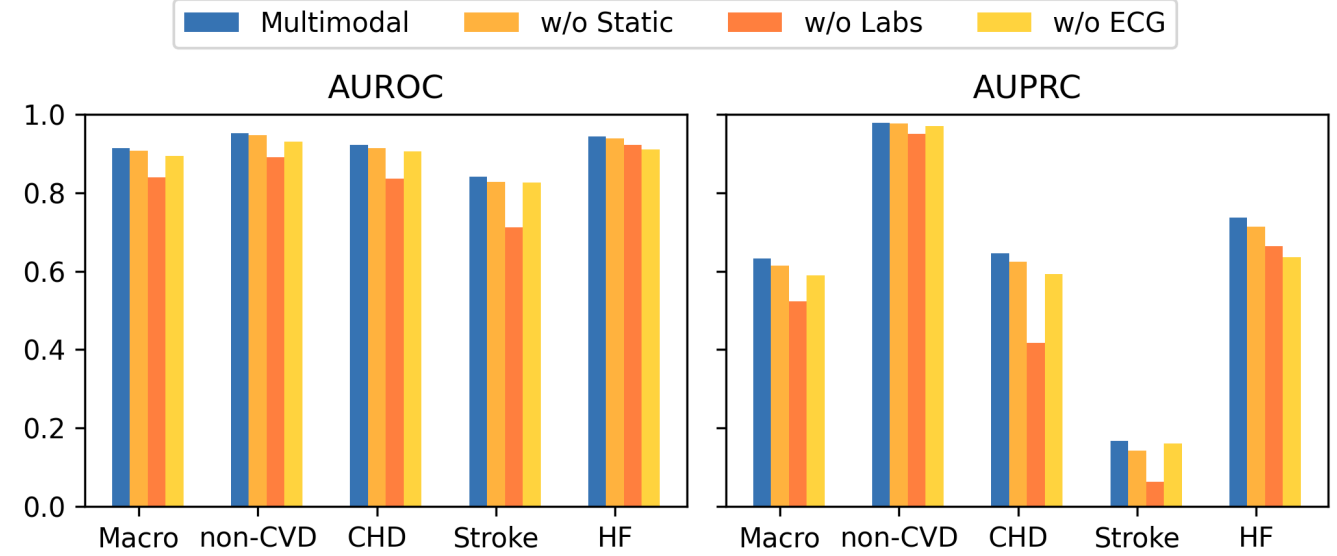

Fig. 5. The performance of multiclass CVD prediction task (Task 1, core subset). Performance (AUROC and AUPRC) across Macro-average and four classes: non-CVD (N=125,987), CHD (coronary artery disease; N=18,445), Stroke (N=4,927), and HF (heart failure; N=21,418). Multimodal integration (Static + Labs + ECG) is compared with ablations removing one modality at a time. Excluding labs most affected Macro, non-CVD, CHD, and Stroke, while excluding ECG most impacted HF, highlights the complementary roles of different modalities.

To evaluate MedM2T's robustness to missing ECG data, we assessed in-hospital mortality prediction (Task 2) under varying levels of data availability. In the full cohort (N=40,167), despite a 53.3% missing rate for ECG records, MedM2T maintained high predictive performance (AUROC 0.868, AUPRC 0.470). This performance is comparable to the results achieved on the ECG-available subset (N=18,750; AUROC 0.876, AUPRC 0.475), suggesting that the model is resilient to missing modalities. Crucially, even with substantial missingness, the multimodal approach outperformed the ECG-excluded baseline (AUROC 0.862, AUPRC 0.455), demonstrating that our architecture successfully leverages available signals to enhance overall prediction.

We further evaluated MedM2T on an external dataset, using the MC-MED dataset [25], [26], from a different institution and a clinically distinct emergency department setting, focusing on acute diagnostic stratification of cardiopulmonary conditions. MC-MED including structured EHR variables, sparse laboratory tests, mixed-resolution vital signs, continuous ECG waveforms, and free-text radiology reports. For free-text radiology reports, we followed the text encoding strategy recommended by HAIM [2] and employed Clinical BERT [27] as a modality-specific encoder. Among unimodal models, while laboratory features achieved the strongest overall performance (**Fig. 6**), particularly for acute myocardial infarction, the predictive utility of other sources is highly task-specific. For instance, static features are the primary driver for acute heart failure or pulmonary edema, whereas radiology text provides greater discriminative value for pulmonary embolism, and vital signs remain essential for pneumonia or sepsis. By integrating all modalities, MedM2T achieves the best macro-averaged performance (AUROC 0.880, AUPRC 0.392) and consistently improves class-wise discrimination.

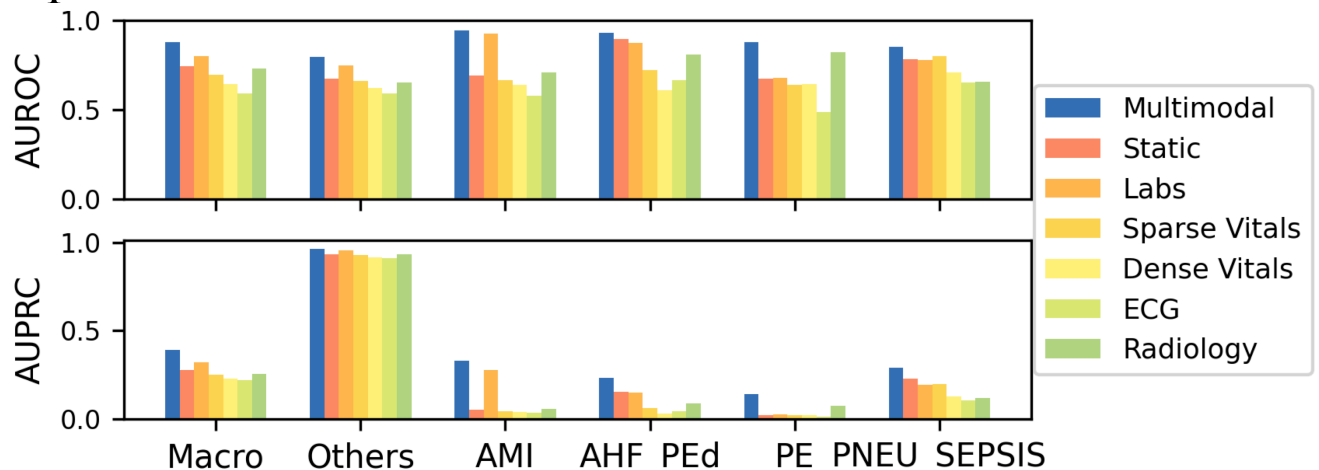

Fig. 6. Multiclass classification performance on the MC-MED external dataset. Class-wise results are grouped by multimodal and unimodal models across five categories: AMI (acute myocardial infarction), AHF_PEd (acute heart failure or pulmonary edema), PE (pulmonary embolism), PNEU_SEPSIS (pneumonia or sepsis), and Others. The multimodal model achieves superior macro-averaged performance and improved discrimination across classes. Additional results are provided in **Supplementary Table III**.

### D. Computational Cost

TABLE VII
GPU MEMORY AND RUNTIME COMPARISON FOR TASK 1

| | | Alloc. Mem. | Res. Mem. | Time |
|---|---|---|---|---|
| | | Training / Inference | | |
| **Multimodal** | **MedM2T** | 2.2 / 0.6 | 6.1 / 1.4 | 68.9 / 26.6 |
| | **Late Fusion** | 1.9 / 0.4 | 5.7 / 1.1 | 63.2 / 25.6 |
| **Unimodal** | **Static** | 0.3 / 0.3 | 0.4 / 0.4 | 0.2 / 0.7 |
| | **Labs** | 0.1 / 0.1 | 0.1 / 0.1 | 6.7 / 1.7 |
| | **ECG** | 1.9 / 0.4 | 5.0 / 1.1 | 59.7 / 22.7 |
| | **ECG-E** | 0.1 / 0.1 | 1.2 / 1.0 | 17.9 / 16.4 |

Alloc. Mem. and Res. Mem. denote average per-batch allocated and reserved GPU memory (GiB), respectively, with batch size = 32; Time denotes average per-batch runtime (ms). ECG-E denotes using pre-computed ECG embeddings instead of raw ECG.

**Table VII** summarizes the Task 1 computational costs, demonstrating that MedM2T introduces only marginal additional GPU memory consumption and runtime relative to late fusion. Detailed analysis identifies the primary computational bottleneck as the hierarchical time-aware ECG encoder processing high-resolution raw signals, rather than the fusion mechanism itself. To mitigate this cost, we use a pre-trained ECG model to pre-compute and store ECG embeddings for reuse (ECG-E). This strategy reduces training-time allocated memory, reserved memory, and runtime by 94%, 76%, and 70%, respectively, and achieves inference-time reductions of 81%, 9%, and 27.7%.

### E. Interpretability Analysis

To analyze the relative importance of each modality and its features, we used Integrated Gradients (IG; Captum) [28], [29] to quantify feature-level contributions to model predictions and to probe cross-modal interactions by attributing both pairwise Bi-Modal Attention outputs and the aggregated components in the cross-modal fusion stage. We analyzed feature- and modality-level contributions for HF prediction in Task 1 based on static and laboratory modalities. The attribution of laboratory features showed a clear temporal trend, with the highest contribution scores concentrated in the first time windows immediately nearing the final ECG measurement (**Fig. 7**). This effect was particularly dominant in markers of renal function and myocardial injury. Decomposition of the decoder revealed comparable contributions from Static (37%), Labs (33%), and their Bi-Modal Attention interaction (30%), with the interaction pathway primarily driven by the Labs branch (86%) (**Supp. Fig. 3**). In another representative true-positive HF case predicted by multimodal ECG, the model attended to arrhythmic cues, including irregular RR intervals and the absence of detectable P-waves (**Supp. Fig. 4**). These factors were concurrently observed across the signal, textual, and feature data. Complementing these feature-level insights, the calibration curves in **Supp. Fig. 5** justify the reliability of the predicted probabilities, supporting the use of these calibrated outputs for clinical alert generation in Tasks 1 and 3.

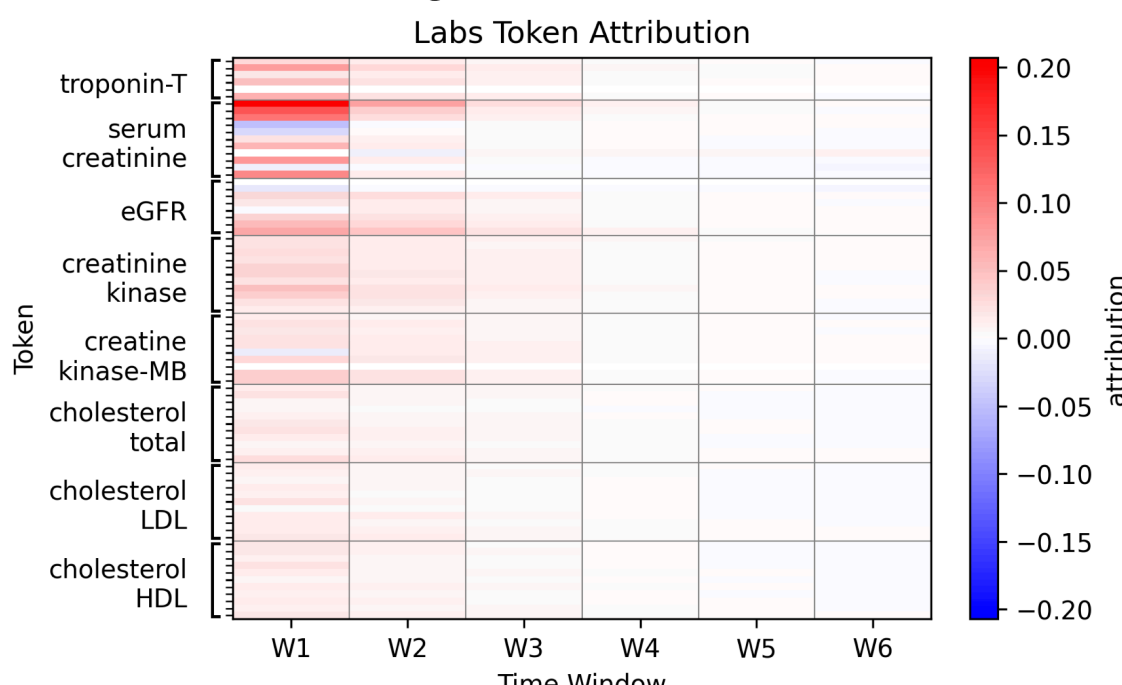

Fig. 7. Laboratory feature attributions for Heart Failure (HF) prediction in Task 1. For each laboratory item, tokens are ordered from top to bottom by decreasing percentile value. The heatmap shows attribution scores across temporal windows relative to the last ECG measurement time $T$. Laboratory features exhibit the strongest attribution near $T$. Specifically, contributions from renal and myocardial biomarkers (creatinine, eGFR, and troponin) reflect the cardiorenal involvement characteristic of HF decompensation, indicating the model's adherence to established clinical pathophysiology.

## IV. DISCUSSION

### *Multimodal Framework*

We proposed MedM2T, a multimodal fusion framework that outperformed existing state-of-the-art frameworks (**Table V**). These frameworks have advanced multimodal integration; however, they often overlook modality-specific characteristics and temporal heterogeneity, relying instead on generic models to process heterogeneous data. MedM2T addresses this gap by leveraging modality-specific encoders tailored to each modality and introduces mechanisms to reduce inter-modality discrepancies, enabling more effective fusion.

The ablation studies show that the pre-training of modality-specific encoders plays the most critical role, primarily by reducing inter-modality discrepancies and improving fusion efficacy for robust multimodal learning. While removing the Bi-Modal Attention yields a statistically significant but smaller performance drop across the full cohorts, its impact becomes more substantial in the ECG-available CICU subset (**Table VI**). In this targeted cohort, Bi-Modal Attention yields notable improvements, suggesting that interaction modeling can be more effective in more homogeneous cohorts or clinically specific subpopulations. In contrast, the smaller gains observed in the full cohorts likely reflect the inherent heterogeneity that attenuates the impact of complex fusion. Finally, the Shared Encoder was introduced to improve representational compatibility across modalities and to facilitate stable interaction learning; however, the ablation results show that the fusion pipeline remains reasonably robust even without this alignment step.

Substituting our encoders into existing multimodal frameworks yielded mixed results, consistently improving performance in MultiModN but not in MultiBench. We hypothesize that MultiBench's design may not offer the requisite flexibility to handle the highly diverse features from our encoders, making it sensitive to the choice of feature extractor and thus preventing consistent performance gains.

The comparative results between MedM2T and HAIM highlight the task-dependent strengths of different modeling methods. For Task 2 (in-hospital mortality), single-modality and ablation analyses consistently indicate that laboratory measurements are the dominant source of predictive information. In this setting, HAIM already achieves strong performance using the labs modality alone (AUROC 0.858, AUPRC 0.470), which largely contributes to its advantage over MedM2T on this task. This result is consistent with prior findings that gradient-boosted tree models are particularly effective for structured tabular data [30], [31]. In Task 2, laboratory measurements are collected within a short temporal window (the first 24 hours after admission) and remain sparse, with an average of only 2.3 observations per observed laboratory variable per patient. Under such conditions, HAIM's simple statistical aggregation suffices to capture clinically relevant threshold effects and non-linear interactions. To summarize, HAIM is well-suited for settings dominated by structured modalities, whereas MedM2T is designed for scenarios with higher modality heterogeneity and more complex representations, such as physiological signals and clinical text. This distinction helps explain why MedM2T shows clearer advantages in Task 1 and Task 3, while HAIM performs better on Task 2.

Ablation experiments on multiclass (**Fig. 5**) show that each modality contributes unique yet complementary information, emphasizing the necessity of multimodal integration for reliable predictions. Moreover, MedM2T shows robustness to a substantial amount of missing data for a modality, yet can effectively leverage that data when it is present.

The external validation on MC-MED (**Fig. 6**) suggests that MedM2T generalizes beyond the MIMIC ecosystem to a different clinical scenario. For instance, radiology reports can be incorporated via a standard text encoder without modifying the overall fusion architecture, supporting applicability across institutions, settings, and modality configurations. The unimodal results further indicate that the most informative modality is diagnosis-dependent, indicating that multimodal integration can improve class-wise discrimination for ED triage in heterogeneous clinical settings.

***Time-Aware Modeling***

To address the unique temporal characteristics of clinical data, we introduced two time-aware modeling strategies. These modalities exhibit markedly different sampling dynamics, motivating modality-specific temporal modeling. **Fig. 8** highlights the temporal heterogeneity between laboratory tests and vital signs by examining their missingness profiles within an hourly window. Laboratory measurements exhibit persistent sparsity over the 24-hour period, reflecting their event-driven and irregular acquisition. In contrast, vital signs exhibit a continuous, regularly sampled pattern, with missingness more often reflecting transient sensor unavailability rather than irregular timing. Motivated by these distinct sampling dynamics, we model laboratory data as sparse time series, while vital signs as continuous physiological signals, and apply continuity-preserving preprocessing. Modeling each modality according to its distinct temporal characteristics enables more effective temporal representation learning in heterogeneous clinical data settings.

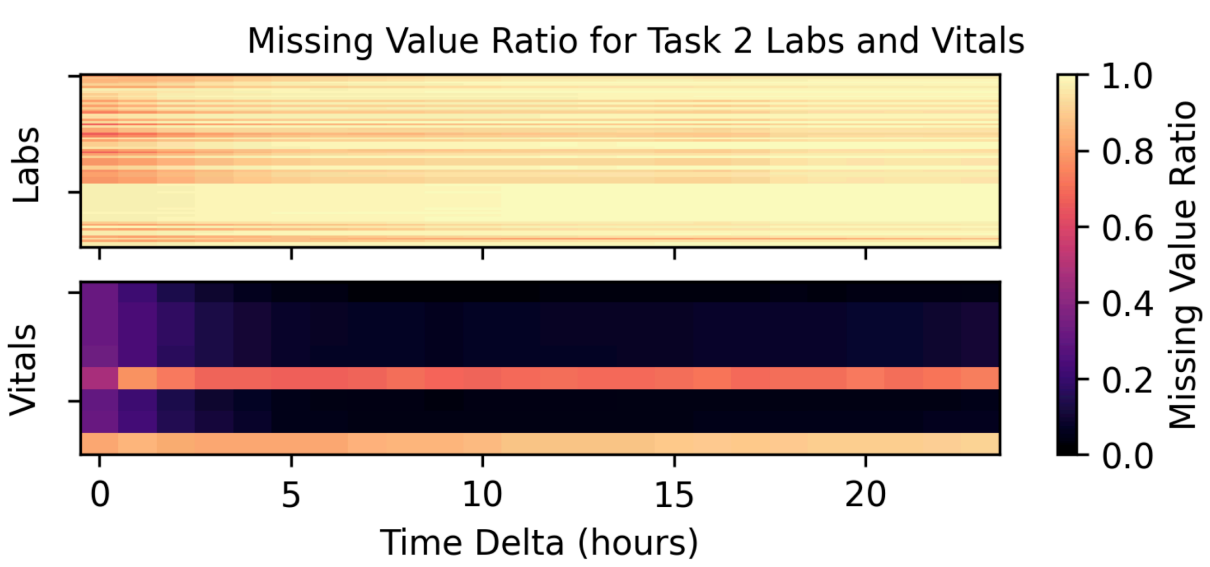

Fig. 8. A comparison of the hourly missingness patterns for laboratory tests and vital signs in Tasks 2 and 3. The heatmap illustrates clear differences in missingness patterns, with laboratory tests persistently sparse over 24 hours and vital signs sampled at more regular intervals.

As illustrated in **Fig. 9,** laboratory records in Task 1 span multiple years and exhibit severe sparsity, with over 80% missing in most intervals and some items completely missing for most patients. Our Sparse Time Series Encoding effectively addresses such challenges. **Table I** shows that our proposed method scales more favorably with increasing temporal span, making it more suitable for practical deployment with long, heterogeneous clinical histories.

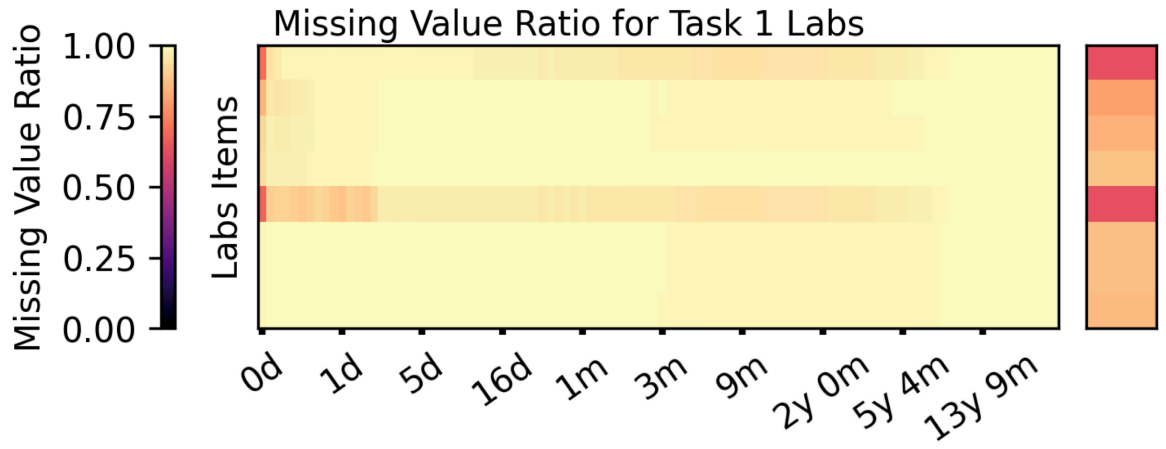

Fig. 9. Missing value ratio for laboratory records in Task 1 across exponentially spaced time intervals. The timeline is divided into 100 intervals based on the absolute time difference from the index date. The heatmap shows that laboratory records are highly sparse, and several lab items are entirely missing across all time points (rightmost column).

For denser time-series data, such as vital signs and frequent ECG records, our Hierarchical Time-Aware Fusion method successfully captures both micro- and macro-level temporal patterns. This approach yielded consistent performance gains across all tasks when applied to dense vital signs data (**Table II**). The method's effectiveness on ECG data was shown to be directly correlated with the availability of longitudinal records. In long-term Task 1, where a substantial portion of the cohort had multiple ECGs (72% with ≥2 ECGs, 43% with ≥5), our method provided a clear performance improvement. In contrast, for the short-term Tasks 2 and 3, where longitudinal ECG data was scarce (only 26% with ≥2 ECGs and 0.5% with ≥5), no statistically significant improvement was observed. This result confirms that our hierarchical approach effectively leverages long-term temporal information when it is available.

The study also confirms the powerful benefit of fusing the multiple data types available from a single ECG event—namely, its raw signal, derived features, and textual notes. This underscores the importance of leveraging the full spectrum of available information, as different data types from the same source provide unique and complementary predictive signals.

Although MedM2T does not explicitly model temporally coupled interactions across modalities, this design avoids enforcing strict temporal alignment in the presence of substantial temporal heterogeneity. Since clinical measurements are often recorded asynchronously, simultaneous observations across modalities are rare. Consequently, enforcing temporal alignment in such cases would substantially increase modality sparsity. Prior studies show that handling temporally heterogeneous and sparse multimodal data often requires additional modeling effort; otherwise, models may over-rely on consistently observed modalities or introduce spurious information through imputation [32], [33] In clinical settings with substantial temporal heterogeneity, alignment-free designs such as MedM2T may offer better generalizability by preserving modality-specific temporal structures while avoiding excessive sparsity.

### *Clinical Applicability: Efficiency and Interpretability*

To assess the practicality of MedM2T for real-world clinical use, we discuss two deployment-critical aspects regarding computational efficiency and interpretability. Efficient execution is essential for institutional deployment, while interpretability provides human-readable evidence that can support clinical trust and decision-making by verifying whether the model's focus aligns with clinical expectations. Such analyses may also offer additional decision support signals and, potentially, facilitate novel insight discovery.

On the efficiency side, our results (**Table VII**) suggest that the proposed architecture, including shared encoders and Bi-Modal Attention, does not introduce prohibitive overhead. Instead, the dominant cost is driven by high-resolution ECG signal encoding rather than the fusion mechanism. This observation motivates a deployment-friendly strategy wherein compact ECG embeddings are pre-computed and stored for subsequent reuse. Because historical physiological recordings are typically fixed in clinical workflows, embedding reuse can reduce repeated signal-level processing and storage demands while preserving MedM2T's ability to model cross-modal temporal interactions.

On the interpretability side, case-level analysis indicates that MedM2T bases its predictions on clinically plausible multimodal evidence, including medication patterns, acute renal/cardiac laboratory dynamics, and rhythm-related ECG cues consistent with known arrhythmias. Beyond individual features, examining modality- and interaction-level contributions can further characterize the relative importance and complementary roles of modalities, providing more structured explanations for clinical review. Together, these findings motivate future work toward an integrated, clinician-facing interpretability interface to improve usability in practice.

### *Limitations*

Limitations of this work should be noted. First, the framework is not designed for interpretability. Although we present representative case studies using post-hoc analyses, the model does not natively provide feature-level explanations of how specific variables interact across modalities and time. Second, Bi-Modal Attention only captures pair-wise interactions, potentially overlooking higher-order multimodal dependencies. Third, temporal relations across modalities are treated independently, which may ignore potential cross-modal interactions within adjacent time segments. Fourth, our evaluation is limited to structured data and ECG signals. Although radiology reports were included in the external validation, other modalities such as medical imaging have not yet been validated. Finally, we employed relatively simple backbone models; while advanced architectures may yield further improvements, our results show the framework is effective even with basic designs.

Future work will focus on enhancing model interpretability, extending MedM2T to additional modalities, and exploring stronger backbone architectures to further improve performance. Most importantly, we aim to capture higher-order and temporally coupled cross-modal interactions. One direction is to extend the hierarchical time-aware fusion mechanism to enable cross-modal interactions at the micro-temporal level, allowing modalities within a fine-grained time window to be jointly modeled while preserving modality-specific temporal resolutions. Additionally, we plan to develop dynamic, time-evolving multimodal fusion strategies in which cross-modal interactions are updated at each time step. This would enable the model to learn how cross-modal relationships evolve and to capture temporal context and causal dependencies across modalities.

## V. Conclusion

In this study, we proposed MedM2T, a flexible and effective multimodal framework that enables a robust fusion of heterogeneous clinical data. By integrating Modality-Specific Encoding with two novel time-aware strategies, MedM2T successfully addresses the heterogeneous temporal characteristics present in clinical data that better align with real-world healthcare scenarios. Furthermore, combining it with Shared Encoding and Bi-Modal Attention, our approach achieves superior performance on long-term (spanning months to years, for CVD prediction) and short-term (within hours, for mortality and LOS prediction) tasks, effectively evaluating both chronic and acute disease dynamics. These tasks also carry direct clinical relevance, with CVD prediction providing early warning, while mortality and LOS serve as critical risk factors in intensive care. We also employ sparse, irregular ECG tests to build an ECG-driven time-aware paradigm that integrates notes, features, and signals for effective use of heterogeneous and non-continuous clinical data. These experiments validated the generalizability and effectiveness of our approach.

## Appendix

The implementation of MedM2T is available at our GitHub repository: https://github.com/DHLab-TSENG/MedM2T.

## A. Dataset

**Supp. Fig. 1** illustrates the population selection process for the CVD dataset used in Task 1. **Supp. Fig. 2** depicts the population selection process for the Mortality dataset, in which 10.04% of patients experienced in-hospital death. The LOS dataset employed the same population as the Mortality dataset, with an average LOS of 4.05 days (SD = 5.19).

Additional details, including the comprehensive dataset description, are provided in the online Supplementary Material (**A: Dataset**).

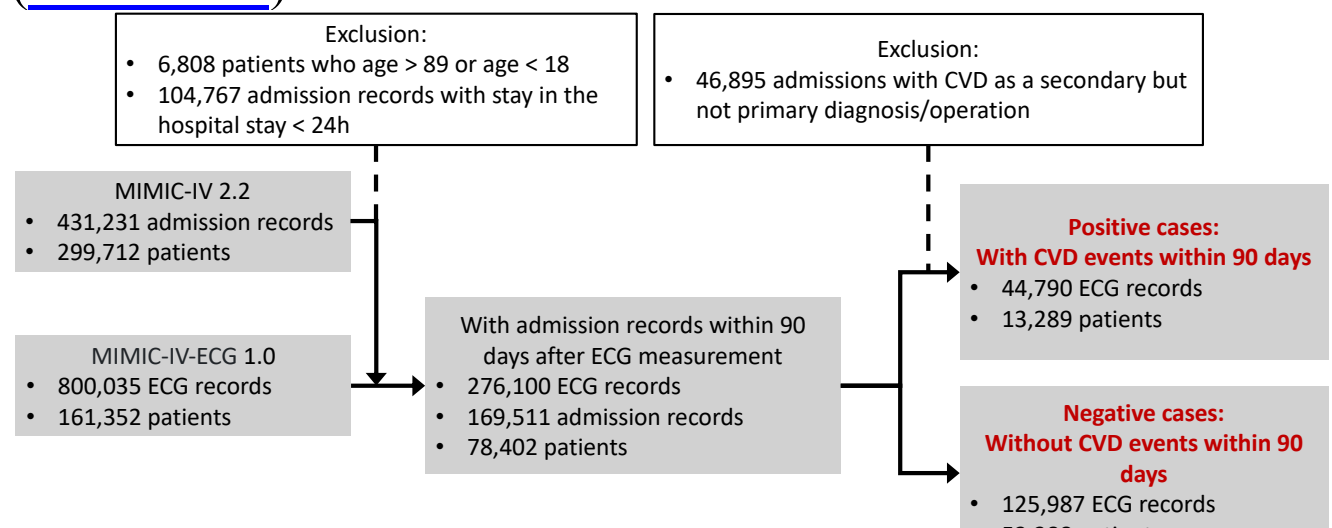

Supplementary Fig. 1. Population selection process for Task 1 (CVD dataset). Focusing on patients with at least one hospitalization occurring within 90 days after an ECG measurement. The final dataset included 44,790 CVD-related and 125,987 non-CVD ECG samples.

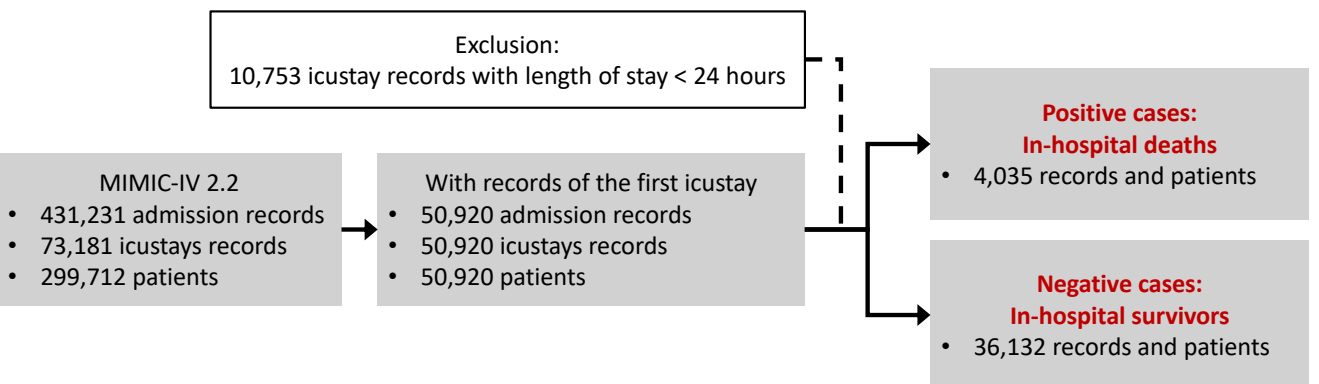

Supplementary Fig. 2. Population selection process for Task 2 (in-hospital mortality). The final cohort included 40,167 first ICU admission records, with 4,035 mortality cases (10.04%).

## B. Experimental Setup

**Hyperparameters of MedM2T:** The detailed hyperparameter settings of MedM2T are provided in the online Supplementary Material (**B.I: Hyperparameters of MedM2T**). **Hyperparameters of compared models:** For MultiBench and MultiModN, most hyperparameters were adopted from their original configurations on the MIMIC datasets, with minor adjustments to hidden dimensions and learning rates. HAIM followed the hyperparameter tuning strategy recommended in the original work. Full details are provided in the online Supplementary Material (**B.II: Hyperparameters of compared models**).

**Sample sizes of tasks:** For unimodal tasks, only subsets of samples with available data in the respective modality were used, whereas multimodal tasks used the entire cohort. Detailed sample sizes and related statistics are provided in the online Supplementary Material (**B.III: Sample sizes of tasks**).

## C. Experimental Results

### *Evaluation of the Multimodal Framework*

**Supp. Table I** presents the results of MedM2T under all multimodal combinations across the three tasks, providing supplementary details corresponding to **Table IV** in the main manuscript. **Supp. Table II** compares the performance of MedM2T with other state-of-the-art multimodal frameworks under unimodal settings across the three tasks, providing supplementary details corresponding to **Table V** in the main manuscript.

Supplementary Table I

| | Task 1 (CVD) | | Task 2 (Mortality) | Task 3 (LOS) |
|---|---|---|---|---|
| | *AUROC ↑* | *AUPRC ↑* | *AUROC / AUPRC ↑* | *MAE / MSE ↓* |
| S+L | 0.893 / 0.924* | 0.590 / 0.651* | 0.839 / 0.397 | 2.49 / 23.13 |
| S+V | - | - | 0.833 / 0.404 | 2.51 / 22.79 |
| S+E | 0.840 / 0.902 | 0.524 / 0.609 | 0.681 / 0.167 | 2.93 / 26.47 |
| L+V | - | - | 0.860 / 0.458 | 2.48 / 22.38 |
| L+E | 0.907 | 0.614 | 0.827 / 0.384 | 2.53 / 23.25 |
| V+E | - | - | 0.815 / 0.379 | 2.50 / 22.97 |
| S+L+V | - | - | 0.862 / 0.455 | 2.40 / 22.26 |
| S+L+E | **0.915 / 0.932*** | **0.632 / 0.670*** | 0.839 / 0.377 | 2.53 / 23.11 |
| S+V+E | - | - | 0.831 / 0.393 | 2.53 / 22.90 |
| L+V+E | - | - | 0.851 / 0.439 | 2.44 / 22.30 |
| S+L+V+E | - | - | **0.868 / 0.470** | **2.33 / 22.21** |

Bold shows the best result; Asterisk (*) denotes using static core subset / extended subset; S, L, V, E denote Static, Labs, Vitals, and ECG.

Supplementary Table II

| | | MedM2T | MultiBench | MultiModN | HAIM |
|---|---|---|---|---|---|
| | | *AUROC / AUPRC ↑* | | | |
| Task 1 (CVD) | S (c) | **0.717** / 0.362 | 0.711 / 0.358 | 0.713 / 0.359 | **0.718 / 0.364** |
| | S (e) | 0.846 / 0.546 | 0.847 / 0.552 | 0.840 / 0.533 | **0.853 / 0.557** |
| | L | **0.870** / 0.568 | 0.852 / 0.542 | 0.829 / 0.511 | 0.864 / **0.574** |
| | E | **0.844 / 0.527** | 0.819 / 0.511 | 0.770 / 0.433 | 0.783 / 0.452 |
| | M (c) | **0.915 / 0.632** | 0.897 / 0.621 | 0.871 / 0.573 | 0.899 / **0.633** |
| | M (e) | **0.932** / 0.670 | 0.916 / 0.649 | 0.889 / 0.593 | 0.923 / **0.680** |
| | | *AUROC / AUPRC ↑* | | | |
| Task 2 (Mortality) | S | **0.678 / 0.170** | 0.671 / 0.164 | 0.677 / 0.166 | 0.672 / 0.165 |
| | L | 0.830 / 0.390 | 0.812 / 0.384 | 0.831 / 0.409 | **0.858 / 0.470** |
| | V | **0.818 / 0.387** | 0.781 / 0.345 | 0.774 / 0.314 | **0.822 / 0.399** |
| | E | **0.710 / 0.191** | **0.691 / 0.183** | **0.679 / 0.179** | **0.688 / 0.176** |
| | M | 0.868 / 0.470 | 0.833 / 0.418 | 0.856 / 0.409 | **0.890 / 0.540** |
| | | *MAE / MSE ↓* | | | |
| Task 3 (LOS) | S | **2.89 / 26.53** | 2.92 / 26.90 | 2.96 / 26.92 | **2.91** / 27.03 |
| | L | **2.51** / 22.82 | 2.96 / 26.91 | 2.95 / 26.99 | 2.53 / **22.47** |
| | V | **2.51 / 22.66** | 2.92 / 26.56 | 2.96 / 26.92 | **2.51 / 22.67** |
| | E | **2.79 / 26.03** | 2.85 / 26.32 | 2.88 / 26.62 | 2.82 / 26.33 |
| | M | **2.33 / 22.21** | 2.93 / 26.82 | 2.95 / 26.92 | 2.43 / **21.89** |

Bold and underlined values represent the best (and statistically comparable, p< 0.05) and second-best results, respectively; MultiBench reports the best multimodal results based on either LR or LRTF; S, L, V, E denote Static, Labs, Vitals, and ECG; M denotes Multimodal; (c), (e) denote core and extended subset.

### *External Validation on the MC-MED Dataset*

**Supp. Table III** reports the results of the external validation on the MC-MED dataset, presenting the performance of each unimodal and the multimodal model. The MC-MED data are derived from emergency department visits at Stanford Health Care. We selected emergency department visits that resulted in admission to inpatient wards or the ICU, and for which both an ECG and a radiology report were available, yielding 18,341 visits.

The prediction target is the primary visit diagnosis, formulated as a five-class classification problem covering (1) acute myocardial infarction (2.44%), (2) acute heart failure or pulmonary edema (2.15%), (3) pulmonary embolism (1.19%), (4) pneumonia or sepsis (6.28%), and (5) others (87.94%) category, with a highly imbalanced class distribution.

Input modalities include static features (demographics, triage measurements, and past medical history), laboratory data comprising 24 clinically relevant tests, vital signs, divided by sampling frequency into sparse (9 variables) and dense (4 variables) measurements, Lead-II ECG waveforms, and free-text radiology reports. Free-text radiology reports are encoded using Clinical BERT. In contrast, the other modalities are encoded using the same modality-specific encoders as in the other tasks, including Sparse Time Series Encoding for laboratory data and sparse vital signs, and Hierarchical Time-aware Fusion for dense vital signs and ECG waveforms.

Supplementary Table III

| | | *AUROC* ↑ | *AUPRC* ↑ |
|---|---|---|---|
| Unimodal | Static | 0.744 | 0.277 |
| | Labs | 0.802 | 0.319 |
| | Sparse Vitals | 0.698 | 0.250 |
| | Dense Vitals | 0.645 | 0.226 |
| | Radiology | 0.730 | 0.253 |
| | ECG | 0.594 | 0.220 |
| Multimodal | Proposed | 0.880 | 0.392 |
| | w/o Pre-trained Encoder | 0.835 | 0.341 |
| | w/o BiModal Attention | 0.878 | 0.392 |
| | w/o Shared Encoder | 0.882 | 0.395 |

### *Interpretability Analysis*

**Supp. Fig. 3** illustrates averaged attributions over true-positive cases for heart failure prediction in Task 1, based on the static (extended) and laboratory modalities. **Supp. Fig. 4** presents a representative true-positive HF case in a multimodal setting, centered on a single ECG measurement to illustrate model behavior at the individual case level.

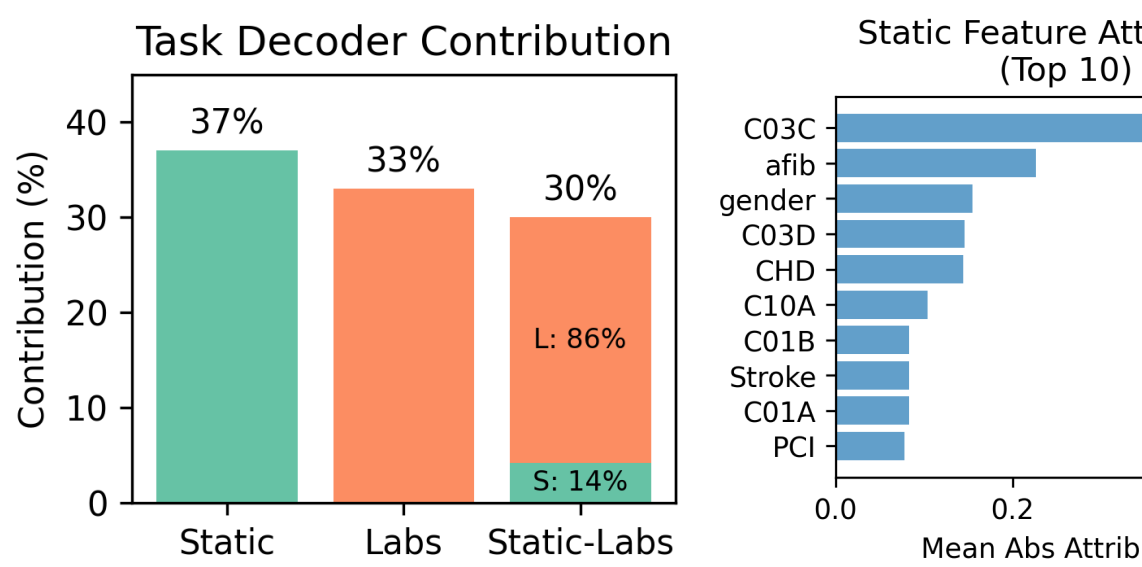

Supplementary Fig. 3. Feature- and modality-level attribution analysis for heart failure prediction in Task 1. *Left*: Decomposition of the final-layer decoder shows comparable contributions from Static, Labs, and their Bi-Modal Attention interaction (Static–Labs). Within the interaction pathway, contributions were primarily driven by the Labs branch (86%), suggesting that acute physiological fluctuations play a dominant role in triggering predictions relative to the static baseline. *Right:* Static attributions reflected treatment and comorbidity signatures, with diuretic use (notably loop diuretics, ATC: C03C) contributing prominently, alongside atrial fibrillation and coronary heart disease (CHD).

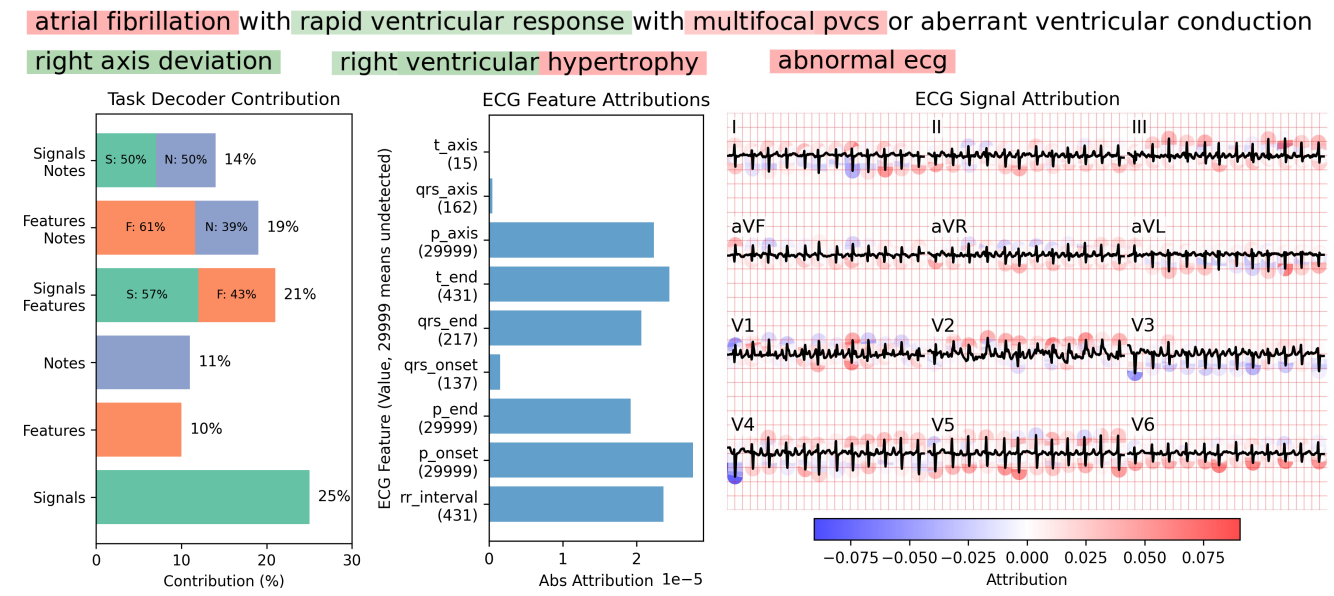

Supplementary Fig. 4. Case-level multimodal attribution analysis for heart failure prediction at a single ECG measurement. Each record comprises clinical text (*Upper*), structured ECG features (*Lower Middle)*, and raw ECG signals (*Lower Right)*. Clinically, HF frequently co-occurs with atrial fibrillation (AF) and premature ventricular complexes. Consistent with this presentation, the model assigned high attribution to rhythm-related cues (e.g., irregular RR intervals) and ventricular depolarization timing features. P-wave-related features were identified as "undetected" (value: 29999), aligning with the absence of P waves during AF. Overall, the attribution patterns demonstrate concordance with established clinical knowledge. *Upper Left*: The decoder decomposition indicates that pairwise Bi-Modal Attention embeddings accounted for more than half of the total contribution, highlighting the importance of cross-modal interaction pathways. Aggregating both unimodal and interaction components, the overall modality contributions were 44.0% for signals, 30.6% for features, and 25.4% for notes.

### *Calibration of Predicted Risk Estimates*

**Supp. Fig. 5** presents the calibration curves for the tasks evaluated, supporting the application of calibrated probabilities for clinical alert generation.

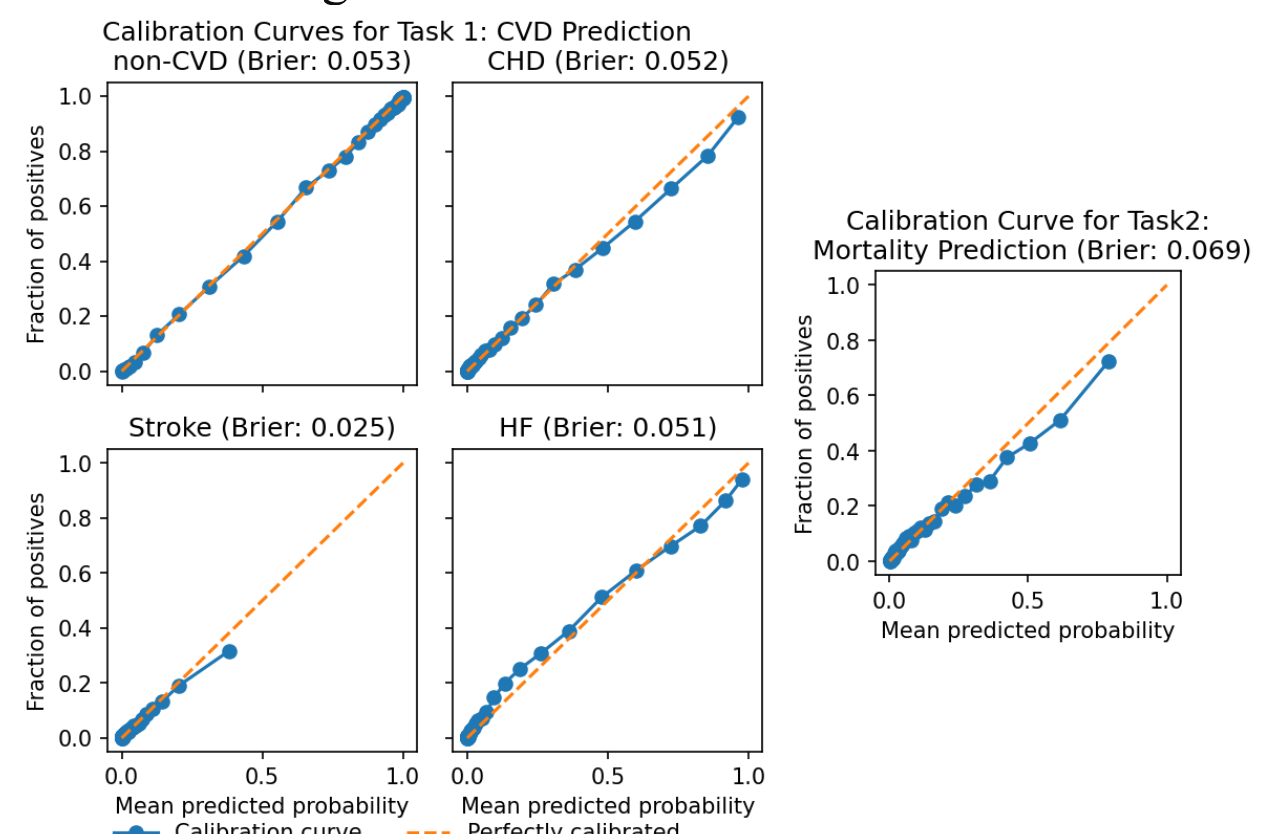

Supplementary Fig. 5. Calibration curves for cardiovascular disease (CVD)–related tasks and in-hospital mortality prediction. Calibration curves are shown for non-CVD, coronary heart disease (CHD), stroke, and heart failure (HF) in Task 1, as well as in-hospital mortality prediction in Task 2. Across all tasks, the predicted risks closely align with the observed event rates, with curves following the ideal diagonal line and low Brier scores ranging from 0.025 to 0.0694. These results indicate reliable absolute risk estimation and support the potential clinical applicability of the proposed model in scenarios requiring calibrated probability outputs.

Additional results, including Task 1 performance across four categories, an extended ablation study on Bi-Modal Attention, and class-wise results on the MC-MED dataset, are provided in the online Supplementary Material (**C: Experimental Results**). The complete online Supplementary Material is available at: https://github.com/DHLab-TSENG/MedM2T/blob/main/SupplementaryMaterial/SupplementaryMaterial.md.